# Neurosymbolic Systems of Perception & Cognition: The Role of Attention

Hugo Latapie [1], Ozkan Kilic [1,*], Kristinn R. Thórisson [2], Pei Wang [3], and Patrick Hammer [4]

[1] *Emerging Technologies & Incubation, Cisco Systems, San Jose, CA, USA*
[2] *Icelandic Institute for Intelligent Machines and Department of Computer Science, Reykjavik University, Reykjavik, Iceland*
[3] *Department of Computer Science, Temple University, Philadelphia, PA, USA*
[4] *Center for Digital Futures, KTH Royal Institute of Technology and Stockholm University, Stockholm, Sweden*

Correspondence*:
Ozkan Kilic
okilic@cisco.com

**ABSTRACT**

A cognitive architecture aimed at cumulative learning must provide the necessary information and control structures to allow agents to learn incrementally and autonomously from their experience. This involves managing an agent's goals as well as continuously relating sensory information to these in its perception-cognition information stack. The more varied the environment of a learning agent is, the more general and flexible must be these mechanisms to handle a wider variety of relevant patterns, tasks, and goal structures. While many researchers agree that information at different levels of abstraction likely differs in its makeup and structure and processing mechanisms, agreement on the particulars of such differences is not generally shared in the research community. A binary processing architecture (often referred to as *System-1* and *System-2)* has been proposed as a model of cognitive processing for low- and high-level information, respectively. We posit that cognition is not binary in this way and that knowledge at *any* level of abstraction involves what we refer to as *neurosymbolic* information, meaning that data at both high and low levels must contain *both* symbolic and subsymbolic information. Further, we argue that the main differentiating factor between the processing of high and low levels of data abstraction can be largely attributed to the nature of the involved attention mechanisms. We describe the key arguments behind this view and review relevant evidence from the literature.

## 1 INTRODUCTION

Cognitive architectures aim to capture the information and control structures necessary to create autonomous learning agents. The sensory modalities of artificially intelligent (AI) agents operating in physical environments must measure relevant information at relatively low levels of detail, commensurate with the agent's intended tasks. Self-supervised learning makes additional requirements on the ability of an agent to dynamically and continuously relate a wide variety of sensory information to high-level goals of tasks. The more general an agent's learning is, the larger a part of its perception-cognition "information stack" must capture the necessary flexibility to accommodate a wide variety of patterns, plans, tasks, and goal structures. Low levels of cognition (close to the perceptual senses) seem to quickly generate and use predictions to generalize across similar problems. This is a key responsibility of a sensory system because low-latency predictions (i.e. those that the agent can act quickly on) are vital for survival in a rapidly changing world. As pointed out by Bengio et al. (2021), natural intelligence has two outstanding skills, compared to Deep Learning: It does not require thousands of samples to learn, and it can cope with out-of-order (OOD) samples. But there is more: Deep Learning does not handle learning after the system leaves the laboratory – i.e. cumulative learning – in part because it does not harbour any means to verify newly acquired information *autonomously*. Such skills require not only perception processes that categorize the sensory data dynamically so that the lower levels can recognize 'familiar' situations by reconfiguring known pieces and trigger higher-level cognition in the case of surprises, but also the reasoning to evaluate the new knowledge that has been thus produced. Whenever high-level cognition solves a new problem, the coordination allows the new knowledge to modify and improve the lower levels for similar future situations, which also means that both systems have access to long-term memory. Architectures addressing both sensory- and planning-levels of cognition are as of yet few and far between.

While general agreement exists in the research community that information at different levels of abstraction likely differs in makeup and structure, agreement on these differences – and thus the particulars of the required architecture and processes involved – is not widely shared. It is sometimes assumed that lower levels of abstraction are subsymbolic and higher levels symbolic, which has led some researchers to the idea that Deep Learning models are analogous to perceptual mechanisms while higher levels involve rule-based reasoning skills due to a symbolic nature, and is the only system that can use language. This binary architectural proposal in some cases be traced back to Kahneman's (2011) approach for a strongly divided architecture (called *System-1* and *System-2*), originally intended to explain results on processing speed differences in humans on various mental tasks. This view has been adopted in some AI research, where 'subsymbolic' processing are classified as System-1 processes, while higher-level and 'symbolic' processing is considered belonging to System-2. According to this view, artificial neural networks, including Deep Learning, are System-1 processes; rule-based systems are System-2.

We posit instead that cognition is not binary in this way at all, and that *any* level of abstraction involves processes operating on what might be called *"neurosymbolic"* knowledge, meaning that data at both high and low levels must accommodate *both* symbolic and subsymbolic information. Further, we argue that the main differentiating factor between the processing of high and low levels of data abstraction can be largely attributed to the nature of the involved *attention mechanisms*.

More than a century ago, James (1890) defined attention as "taking possession by the mind, in clear and vivid form, of one out of what may seem several simultaneously possible objects or trains of thought...It implies withdrawal from some things in order to deal effectively with others." Attention is the fundamental currency of intelligence. Low-level cognition like perception is characterized by a relatively high-speed, non-conscious attention, while higher-level cognition seems more "single-threaded", and relatively slower. When people introspect, our conscious threads of attention seem to consist primarily of the latter, while much of our low-level perceptions are unconscious and under the control of autonomous attention mechanisms (see Koch and Tsuchiya, 2006; Marchetti, 2011; Sumner et al., 2006 for evidence and discussion about decoupling attention from consciousness). Low-level perception and cognitive operations may reflect autonomous access to long-term memory through non-conscious attention mechanisms, while higher-level operation may involve the recruitment of deliberate (introspectively-accessible) cognitive control, working memory, and focused attention (Papaioannou et al., 2021).

## 2 ATTENTION'S ROLE IN COGNITION

High-level (abstract) and low-level (perceptual/concrete) cognition work in coordination, not competition, as one would expect after millions of years of evolution. As demonstrated in numerous experiments, and despite some researchers' claims (see e.g. Evans and Elqayam, 2007; Keren, 2013; Monteiro and Norman, 2013), high-level cognition is not superior to low-level cognition and perception; rather, the two levels cooperate to optimize resource utilization.

Through the evolution of the human brain, it seems that language-based conceptual representations replaced sensory-based compositional concepts, explaining the slower reaction times in humans than other mammals, e.g., chimpanzees (Martin et al., 2014). However, this replacement pushed the boundaries of humans' higher-level cognition by allowing complex propositional representations and mental simulations. While animals do not have the propositional property of language, researchers have found some recursion in birdsong (Gentner et al., 2006) and in syntax among bonobos (Clay and Zuberbuhler, 2011). Moreover, Camp (2009) found evidence that some animals think in compositional representational systems. In other words, animals seem to lack propositional thought, but they have compositional conceptual thought, which is mostly based on integrated multisensory data. Since animals appear to have symbol-like mental representations, these findings indicate that their lower levels are neurosymbolic.

Two separate issues in the System-1/System-2 discussion are often confused: (1) Knowledge representation and (2) information processing. The first is the (by now, familiar) 'symbolic↔subsymbolic' distinction, while the second involves the 'automatic↔controlled' distinction. Not only are these two distinctly different, they are also not always aligned; while subsymbolic knowledge may be more often processed 'automatically' and symbolic knowledge seem generally more accessible through voluntary control and introspection, this mapping cannot be taken as given. A classic example is skill learning like riding a bike, which starts as a controlled process, and gradually becomes automatic with increased training. On the whole this process is largely "subsymbolic," with hardly anything but the top-level goal introspectively accessible to the learner of bicycle-riding ("I want to ride this bicycle without falling").

Among the processes of key importance in skill learning, to continue with that example, is attention; a major cognitive difference between a skilled bike rider and a learner of bike-riding is what they pay attention to: The knowledgeable rider pays keen attention to the tilt angle and speed of the bicycle, responding by changing the angle of the steering wheel dynamically, in a non-linear relationship. Capable as they may already be of turning the front wheel to any desired angle, a learner is prone to fall over in large part because they don't know what to pay attention to. This is why one of the few obviously useful tips that a teacher of bicycle-riding can give a learner is to "always turn the front wheel in the direction you are falling."

Kahneman (1973) sees attention as a pool of resources which allows different process to share cognitive capabilities and posits that there are two "modes" of thinking: 'System-1' is a fast, intrinsic, autonomous, emotional, parallel, and ideally, experience-based, and 'System-2' being slower, deliberate, conscious, and serial (Kahneman, 2011). For example, driving a car on an empty road (with no unexpected events), recognizing your mother's voice, and calculating 2+2, mostly involve System-1, whereas counting the number of people with eyeglasses in a meeting, recalling and dialing your significant other's phone number, calculating 13x17, and filling out a tax form depend on System-2. Kahneman's System-1 is good at making quick predictions because it constantly models similar situations based on experience. It should be noted that "experience" in this context is related to the transfer of learning, which heavily involves higher-level cognition, and should thus be part of Kahneman's System-2. Learning achieved in conceptual symbolic space can be projected to subsymbolic space. In other words, since symbolic and subsymbolic spaces are in constant interaction, acquired knowledge in symbolic space has correspondences in subsymbolic space allowing System-1 to start quickly using the projections of the knowledge even based on System-2 experience.

Several fMRI studies support the idea that sensory-specific areas, such as thalamus, may be involved in multi-sensory stimulus integrations (Miller and D'Esposito, 2005; Noesselt et al., 2007; Werner and Noppeney, 2010), which are symbolic representations in nature. Sensory-specific brain regions are considered to be networks specialized in subsymbolic data that originates from the outside world and different body parts. Thalamo-cortical oscillation is known as a synchronization mechanism or temporal binding between different cortical regions (Llinas, 2002). However, recent evidence shows that the thalamus, previously assumed to be responsible only for relaying sensory impulses from body receptors to the cerebral cortex, can actually integrate these low-level impulses (Tyll et al., 2011; Sampathkumar et al., 2021). In other words, in the thalamus there are sensory-based integrations, and they are essential in sustaining cortical cognitive functions.

Wolff and Vann (2019) use the term "cognitive thalamus" as a gateway to mental representations due to the recent findings that support the idea that thalamocortical and corticothalamic pathways may play complementary but dissociable cognitive roles (see Bolkan et al., 2017; Alcaraz et al., 2018). More specifically, the thalamocortical pathway (the fibers connecting thalamus to cortex region) can create and save task-related representations, not just purely sensory information, and this pathway is essential for updating cortical representations. Similarly, corticothalamic pathways seem to have two major functions: directing cognitive resources (focused attention) and contributing to learning. In a way, the thalamocortical pathway defines the world for the cortex, and the corticothalamic pathway uses attention to tell thalamus what the cortex needs from it to focus. Furthermore, a growing body of evidence shows that the thalamus plays a role in cognitive dysfunction, such as schizophrenia (Anticevic et al.,2014), Down's syndrome (Perry et al., 2018), drug addiction (Balleine et al., 2015), and ADHD (Hua et al., 2021). These discoveries support other recent findings about the role of the thalamus in cognition via the thalamocortical loop. The thalamus, a structure proficient in using and integrating subsymbolic data actively, describes the world for the cortex by contributing to the symbolic representations in it. On the other hand, the cortex uses attention to direct resources to refresh its symbolic representations from the subsymbolic space.

In Non-Axiomatic Reasoning System, attention plays a similar role, that is resource allocation in terms of inference steps to carry out, whereby inference can compose new representation from existing components, and can update the strength of existing relationships via revision. This also leads to a refreshing of representations in a certain sense, as the system will utilize representations which are more reliable and switch to alternatives if some of them turn out to be unrealiable.

And the Auto-catalytic Endogenous Reflective Architecture (AERA) models attention as system-wide control of computational/cognitive resources (cf. Nivel et al. 2015, Helgason et al. 2013). Studies on human-multitasking have shown that a degree of parallelism among multiple tasks is more likely if the tasks involve different data modalities, such as linguistic and tactile. Low-level attention continuously monitors both mind and the outside world and assesses situations with little or no effort, through its access to long-term memory and the sensory information. Surprises and threats and detected early in the perceptual stream, while plans and questions are handled at higher levels of abstraction, triggering higher levels of processing, which also provide a top-down control of attention and reasoning.

## 3    A NEUROSYMBOLIC ARCHITECTURE AS SYSTEMS OF THINKING

The idea of combining symbolic and sub-symbolic approaches, also known as the neurosymbolic approach, is not new. Many researchers are working on integrated neural-symbolic systems which translate symbolic knowledge into neural networks. Because symbols, relations, and rules should have counterparts in the sub-symbolic space, a compiler for neural networks is needed. Moreover, the neurosymbolic network needs a symbol manipulation that also supports preservation of the structural relations between the two systems without losing the correspondences. Currently, Deep Learning and related machine learning methods are primarily subsymbolic. Meanwhile, rule-based systems and related reasoning systems are usually strictly symbolic. However, it is possible to have a Deep Learning model that demonstrates symbolic cognition, which entails the transformation of symbolic representations into subsymbolic ML/DL/statistical models. One of the costs associated with this transformation is the loss of the underlying causal model which may have existed in the symbolic representation. Current subsymbolic representations are correlation models, where spurious anti-knowledge is representationally indistinguishable from useful correlations and causal knowledge. There is an ongoing interest in bringing symbolic and abstract thinking to Deep Learning, which could enable more powerful kinds of learning. Graph neural networks with distinct nodes (Kipf et al. 2018; Van Steenkiste et al., 2018), transformers with discrete positional elements (Vaswani et al., 2017), and modular models with bandwidth (Goyal and Bengio, 2020) are examples of attempts in this direction. Liu et al. (2021) summarize the advantages of having discrete values (symbols) in a Deep Learning architecture. First, using symbols allows a language for inter-modular interaction and learning, whereby the meaning of symbols is not innate but determined by the relationships with others (as in Semiotics). Second, it allows reusing previously learned symbols in unseen or out-of-order situations, by reinterpreting them in a way suitable to the situation. Discretization in Deep Learning may provide systematic generalization (recombining existing concepts) but it is currently not very successful (Lake and Baroni, 2018).

Current hybrid approaches attempt to combine symbolic and subsymbolic models to compensate for each other's drawbacks. However, the authors believe that there is a need for a metamodel which will accommodate hierarchical knowledge representations. Latapie et al. (2021) proposed such a model inspired by Korzybski's (1994) idea about levels of abstraction. Their model promotes cognitive synergy and metalearning, which refer to the use of different computational techniques and AGI approaches, e.g., probabilistic programming, machine learning/Deep Learning, AERA (Thórisson, 2020, Nivel et al. 2013), OpenNARS[1] (Wang, 2006; Wang, 2010) to enrich its knowledge and address combinatorial explosion issues. The current paper extends the metamodel as a neurosymbolic architecture as in Figure 1.

In this metamodel, the levels of abstractions are marked with L. L0 is the closest to the raw data collected from various sensors. L1 contains the links between raw data and higher level abstractions. L2 corresponds to the highest integrated levels of abstraction learned through statistical learning, reasoning, and such. The layer L2 can have an infinite number of sub-layers since any level of abstraction in L2 can have metadata existing at an even higher level of abstraction. L* holds the high-level goals and motivations, such as self-monitoring, self-adjusting, self-repair, and the like. Similar to the previous version, the neurosymbolic metamodel is based on the assumption of insufficient knowledge and resources (Wang, 2005). The symbolic piece of the metamodel can be thought of as a knowledge graph with some additional structure that includes both a formalized means of handling anti-symmetric and symmetric relations, as well as a model of abstraction. The regions in the subsymbolic piece of the metamodel are mapped to the nodes in the symbolic system in L1. In this approach, the symbolic representations are always refreshed in a bottom-up manner.

---

[1] https://github.com/opennars/opennars — *last accessed on Oct 20th, 2021.*

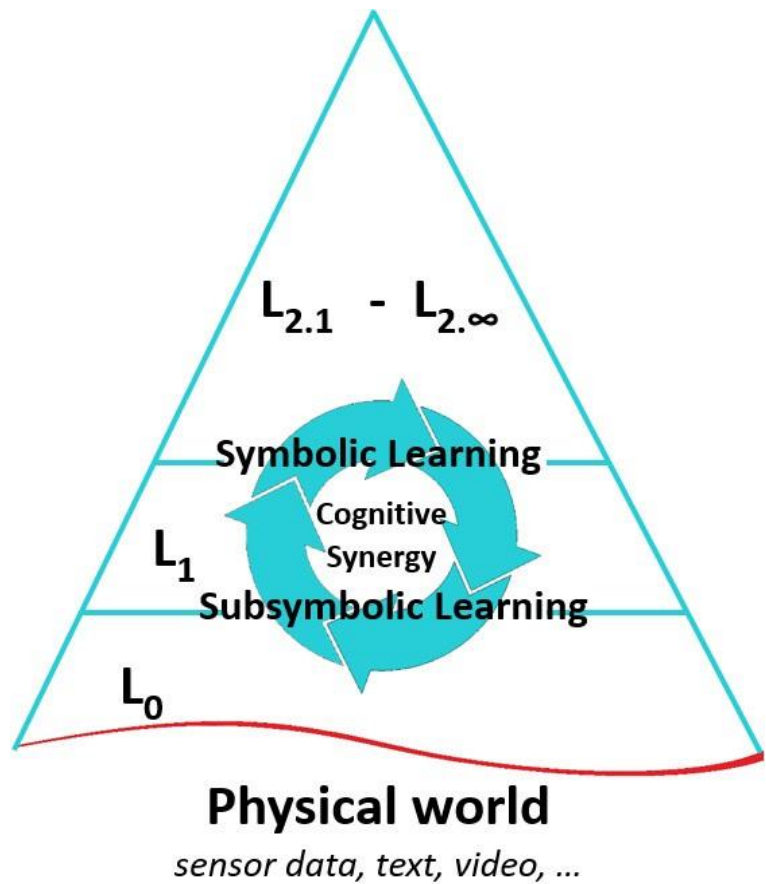

**Figure 1.** Neurosymbolic Metamodel and Framework for Artificial General Intelligence

Depending on the system's goal or subgoals, the metamodel can be readily partitioned into subgraphs using the hierarchical abstraction substructure associated with the current focus of attention. This partitioning mechanism is crucial to manage combinatorial explosion issues while enabling multiple reasoners to operate in parallel. Each partition can trigger a sub focus of attention (sFoA), which requests subsymbolic data from System-1 or some answers from System-2. The bottom-up refreshing and the neurosymbolic mapping between regions and symbols allow the metamodel to benefit from different computational techniques (e.g., probabilistic programming, Machine Learning/Deep Learning and such) to enrich its knowledge and benefit from the 'blessing of dimensionality' (cf. Gorban, 2018), also referred to as 'cognitive synergy.'

A precursor to the metamodel as a neurosymbolic approach was first used in Hammer et al. (2019). This version was the first commercial implementation of a neurosymbolic AGI approach in the smart city domain. Later, the need for use of the levels of abstraction in the metamodel became mandatory due to the combinatorial explosion issue. In other words, structural knowledge representation with the levels of abstraction became very important for partitioning the problem, process subsymbolic or symbolic information for each sub problem (focus of attention, FoA), and then combine the symbolic results in the metamodel. The metamodel with the level of abstraction was actually achieved fully in the retail domain (see Latapie et al., 2021 for details). The flow of the retail use case with the metamodel is shown in Figure 2.

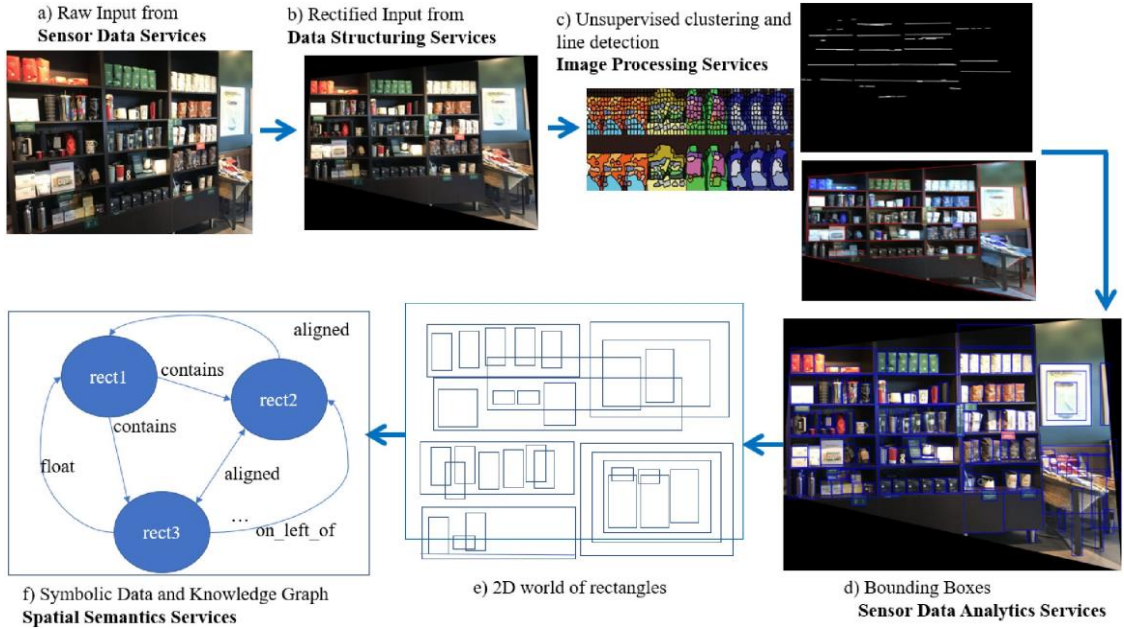

**Figure 2.** Flow of Retail Use Case for Metamodel (from Latapie et al., 2021)

The example for the levels of abstraction using the results of the retail use case is shown in Figure 3.

Latapie et al. (2021) emphasized that no Deep Learning model was trained with product or shelf images for the retail use case. The system used for the retail use case is solely based on representing the subsymbolic information in a world of bounding boxes with spatial semantics. The authors tested the metamodel in 4 different settings with and without the FoA and reported the results as in Table 1.

**Table 1.** Experimental results from Retail Use Case using Metamodel

| Category | without FoA (%) | | | with FoA (%) | | |
|---|---|---|---|---|---|---|
| | precision | recall | f1-score | precision | recall | f1-score |
| *product* | 80.70 | 29.32 | 52.88 | 96.36 | 99.07 | 97.70 |
| *shelf* | 8.82 | 18.75 | 12.00 | 82.35 | 87.50 | 88.85 |
| *other* | 36.61 | 89.66 | 52.00 | 96.00 | 82.76 | 88.89 |
| **overall accuracy** | 46.30 (min/max: 30.13/84.65) | | | 94.73 (min/max: 88.10/100.00) | | |

Another use case for the metamodel is the processing of more than 200,000 time series with a total of more than 30 million individual data points. The time series are network telemetry data. For this use case, there are only two underlying assumptions: The first assumption is that the time series or a subset of them is at least weakly-related, such as time series from computer network devices. The second assumption is that when a number of time series simultaneously change their behaviors, it might indicate that an event-of-interest has happened. For detecting anomalies and finding regime change locations, Matrix Profile algorithms are used (see Yeh et al., 2016; Gharhabi et al. 2017 for Matrix Profile and Semantic Segmentation). Similar to the retail use case, millions of sensory data points are reduced to a much smaller number of events based on the semantic segmentation points. These points are used to form a histogram of regime changes as shown in Figure 4. The large *spikes* in the histogram are identified as the *candidate events-of-interest*. Then the metamodel creates a descriptive model for all time series, which allows system to downsize millions of data points into a few thousand structural actionable and explainable knowledge. To test the metamodel with time series, we first use a subset of the Cisco Open Telemetry Data Set.[2] After being able to identify the anomalies in the data set, we create our own data sets similar to the Open Telemetry Data. For this purpose, 30 computer network events, such as memory leak, transceiver pull, port flap, port shut down, and such, are injected to a physical computer network. The system is able to identify 100% of the events with a maximum of 1 minute delay. For example, Figure 4 represents the histogram of regime changes for a *port shut down* event, which is injected at the 50th timestamp. Since the sampling rate is 6 seconds, one minute later (which is at the 60th timestamp) the system detects a spike as an event-of-interest. It can take time for a single incident to display a cascading effect on multiple devices. When the injection ends at the 100th timestamp, another spike is observed within 10 timestamps, which represents a recovery behavior for the network. It should be noted that not all events necessarily mean an error has happened. Some usual activities in the network, e.g., a usual firmware update on multiple devices as events-of-no-interest, are also captured by the metamodel. The metamodel learns to classify such activities either by observing the network. Although the time series processing using the metamodel does not require any knowledge of computer networking, it can easily incorporate such features extracted by networking-specific modules, e.g., Cisco Joy, [3] or ingest some expert knowledge defined in the symbolic world, specifically at the 2nd level of abstraction This neurosymbolic approach with the metamodel can quickly reduce the sensory data into knowledge, reason on this knowledge, and notify the network operators for remediation or trigger a self-healing protocol.

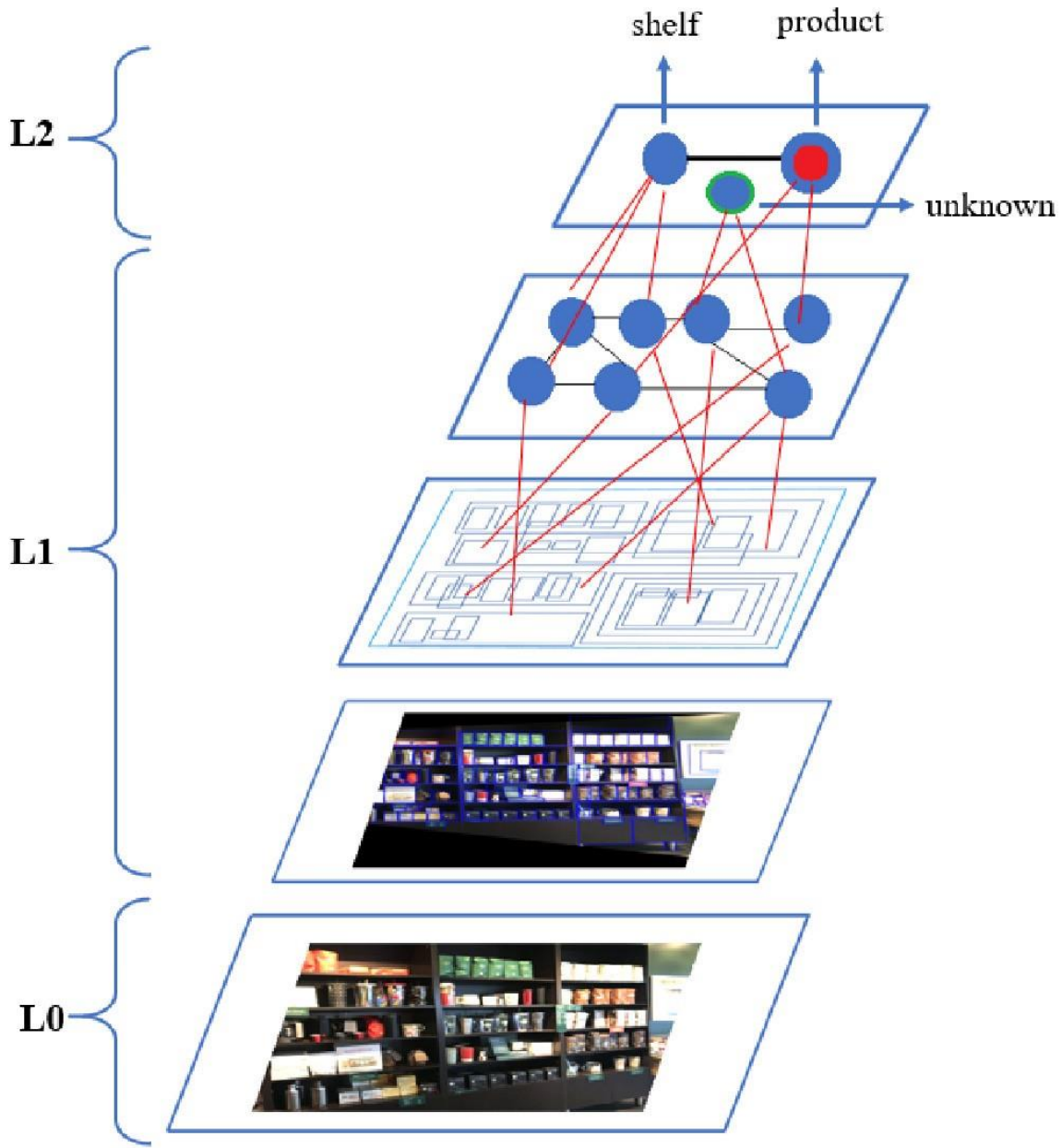

**Figure 3.** Levels of Abstraction for Retail Use Case (from Latapie et al., 2021)

---

[2] https://github.com/cisco-ie/telemetry

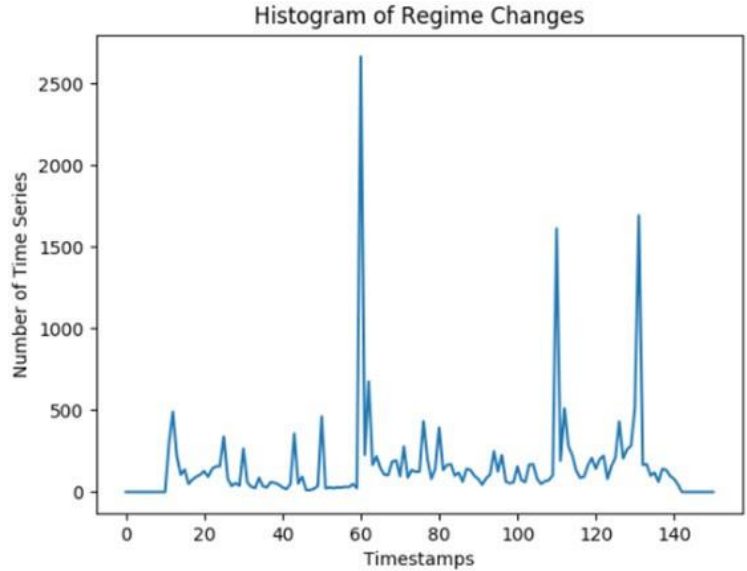

**Figure 4.** A Histogram of Regime Changes from Network Telemetry Data (A *port shut down* event started at the 50th timestamp and ended at the 100th)

## 4 DISCUSSIONS AND CONCLUSION

The metamodel promotes cognitive synergy while preserving level of abstraction, symmetric and anti-symmetric properties of knowledge and using a bottom-up approach to refresh System-2 symbols from System-1 data integration (see Latapie et al., 2021 for details). Moreover, System-1 provides rapid responses to the outside world and activates System-2 in case of a surprise such as an emergency or other significant event that requires further analysis and potential action. System-2 uses conscious attention to request subsymbolic knowledge and sensory data from System-1, to be integrated into the levels of abstraction inspired from Korzybski's work. Korzybski's two major works (Korzybski, 1921; Korzybski, 1994) emphasize the importance of bottom-up knowledge. The corticothalamic and thalamocortical connections play different but complementary roles.

---

[3] https://github.com/cisco/joy

A balanced interplay between System-1 and System-2 is important. System-1's innate role is to ensure the many faceted health of the organism. System-2 is ideally used to help humans better contend with surprises, threats, complex situations, important goals, and achieve higher levels in Maslow's hierarchy of needs. If it dominates the attention of the organism, System-2 can become a source of distress, subverting System-1's ability to achieve health, happiness, and homeostasis. Modern civilization has dramatically advanced our ability to deal with ever higher degrees of abstraction. However, in some cases the cost has been very high: Corticothalamic integration has reversed. In other words, the cortex projects its abstractions, e.g., fears and imaginary stressors, onto the thalamus and hence the body. From an AI systems perspective, current Deep/Machine Learning (including System-2 deep learning) have the opposite problem: Causal modeling and advanced reasoning are being solved in System 1, leveraging statistical models which can be seen as an inversion of proper thalamocortical integration.

## 5 CONCLUSIONS

While not conclusive, findings about natural intelligence from psychology, neuroscience, cognitive science, and animal cognition imply that both low-level perceptual knowledge and higher-level more abstract knowledge may be neurosymbolic. The difference between high and low levels of abstraction may be that lower levels involve a greater amount of unconscious (automatic) processing and attention, while higher levels are introspectable to a greater extent (in humans, at least) and involve conscious (i.e. steerable) attention. The neurosymbolic metamodel and framework introduced in this paper for artificial general intelligence is based on these findings, and the nature of the distinction between both systems will be subject to further research. One may ask whether artificial intelligence needs to mimic natural intelligence as a key performance indicator. The answer is yes and no. No, because natural intelligence, a result of billions of years of evolution, is full of imperfections and mistakes. Yes, because it is the best way known to help organisms survive for countless generations.

Both natural and artificial intelligences can exhibit astounding generalizability, performance, ability to learn, and other important adaptive behaviors when symbolic originating attention and sub-symbolic originating attention are properly handled. Allowing one system of attention to dominate, or inverting the natural order (e.g. reasoning in the subsymbolic space or projecting symbolic space stressors into the subsymbolic space) may lead to suboptimal results for engineered systems, individuals, and societies.

## AUTHOR CONTRIBUTIONS

H. Latapie and O. Kilic conceived of the presented idea. H. Latapie designed the framework and the experiments. H. Latapie and O. Kilic implemented the framework. O. Kilic ran the tests and collected data. K. R. Thórisson, P. Wang and P. Hammer contributed to the theoretical framework. All authors contributed to the writing of this manuscript and approved the final version.

## ACKNOWLEDGMENTS

The authors would like to thank all anonymous reviewers and colleagues for their comments.

## REFERENCES

(2015). Anytime bounded rationality. In *E. Nivel and K. R. Thórisson and B. Steunebrink and J. Schmidhüber*. 121–130

Alcaraz, F., Fresno, V., Marchand, A. R., Kremer, E. J., Coutureau, E., and Wolff, M. (2018). Thalamocortical and corticothalamic pathways differentially contribute to goal-directed behaviors in the rat. *eLife* 7, e32517. doi:10.7554/eLife.32517

Anticevic, A., Cole, M. W., Repovs, G., Murray, J. D., Brumbaugh, M. S., Savic, A. M. W. A., et al. (2014). Characterizing thalamo-cortical disturbances in schizophrenia and bipolar illness. *Cerebral Cortex* 24, 3116–3130. doi:10.1093/cercor/bht165pmid:23825317

Balleine, B. W. and Leung, R. W. M. B. K. (2015). Thalamocortical integration of instrumental learning and performance and their disintegration in addiction. *Brain Research* 1628, 104–116. doi:10.1016/j.brainres.2014.12.023

Bengio, Y., Lecun, Y., and Hinton, G. (2021). Deep learning for ai. *Communications of the ACM* 64, 58–65

Bolkan, S. S., Stujenske, J. M., Parnaudeau, S., Spellman, T. J., Rauffenbart, C., Abbas, A. I., et al. (2017). Thalamic projections sustain prefrontal activity during working memory maintenance. *Nature Neuroscience* 20, 987–996

Camp, E. (2009). Language of baboon thought. In *The Philosophy of Animal Minds*, ed. R. W. Lurz (Cambridge: Cambridge University). 108–127

Clay, Z. and Zuberbühler, K. (2011). Bonobos extract meaning from call sequences. *PLoS ONE* 6, e18786

Evans, J. S. B. and Elqayam, S. (2007). Dual-processing explains base-rate neglect, but which dual-process theory and how? *Behavior and Brain Science* 30, 261–262. doi:10.1017/S0140525X07001720

Gentner, T. Q., Fenn, K. M., Margoliash, D., and Nusbaum, H. C. (2006). Recursive syntactic pattern learning by songbirds. *Nature2006* 440, 1204–1207

Gharghabi, S., Ding, Y., Yeh, C.-C. M., Kamgar, K., Ulanova, L., and Keogh, E. (2017). Matrix profile viii: Domain agnostic online semantic segmentation at superhuman performance levels. In *2017 IEEE International Conference on Data Mining (ICDM)*. 117–126. doi:10.1109/ICDM.2017.21

Gorban, A. N. and Tyukin, I. Y. (2018). Blessing of dimensionality: mathematical foundations of the statistical physics of data. *Philosophical Transactions of the of the Royal Society Mathematical Physical and Engineering Sciences* 440, 1204–1207

Goyal, A. and Bengio, Y. (2020). Inductive biases for deep learning of higher-level cognition. *arXiv preprint*

Hammer, P., Lofthouse, T., Fenoglio, E., and Latapie, H. (2019). A reasoning based model for anomaly detection in the smartcity domain. In *Proc. of AGI-2013*

Helgason, H. P., Thórisson, K. R., Garrett, D., and Nivel, E. (2013). Towards a general attention mechanism for embedded intelligent systems. *International Journal of Computer Science and Artificial Intelligence* 4, 1–7

Hua, M., Chen, Y., Chen, M., Huang, K., Hsu, J., Bai, Y., et al. (2021). Network-specific corticothalamic dysconnection in attention-deficit hyperactivity disorder. *Journal of Developmental and Behavioral Pediatrics* 42, 122–127. doi:10.1097/DBP.0000000000000875

James, W. (1890). *The Principles of Psychology, Vol. 2* (NY: Dover Publication)

Kahneman, D. (1973). *Attention and Effort* (NJ: Prentice-Hall)

Kahneman, D. (2011). *Thinking, fast and slow* (NY: Farrar, Straus and Giroux)

Keren, G. (2013). A tale of two systems: A scientific advance or a theoretical stone soup? commentary on evans stanovic. *Perspectives on Psychological Science* 8, 257–262

Kipf, T., Fetaya, E., Wang, K. C., Welling, M., and Zemel, R. (2018). Neural relational inference for interacting systems. *arXiv preprint*

Koch, C. and Tsuchiya, N. (2006). Attention and consciousness: two distinct brain processes. *Trends Cognitive Science* 11, 16–22. doi:10.1016/j.tics.2006.10.012

Korzybski, A. (1921). *Manhood Of Humanity, The Science and Art of Human Engineering* (NY: E. P. Dutton and Company)

Korzybski, A. (1994). *Science and Sanity: An Introduction to Non-Aristotelian Systems, 5th edn,* (NY: Institute of General Semantics)

Lake, B. and Baroni, M. (2018). Generalization without systematicity: On the compositional skills of sequence-to-sequence recurrent networks. In *Proc. of International Conference on Machine Learning* (PMLR), 2873–2882

Latapie, H., Liu, O. K. G., Kompella, R., Lawrence, A., Sun, Y., Srinivasa, J., et al. (2021). A metamodel and framework for artificial general intelligence from theory to practice. *Journal of Artificial Intelligence and Consciousness* 8, 205–227. doi:10.1142/S2705078521500119

Liu, D., Lamb, A., Kawaguchi, K., Goyal, A., Mozer, C. S. M. C., and Bengio, Y. (2021). Discrete-valued neural communication. *arXiv preprint*

Llinas, R. R. (2002). Thalamocortical assemblies: How ion channels, single neurons and large-scale networks organize sleep oscillations. In *Thalamus and Related Systems*, eds. A. Destexhe and T. J. Sejnowski (Oxford: Oxford University). 87–88

Marchetti, M. (2011). Against the view that consciousness and attention are fully dissociable. *Frontiers in Psychology* 3. doi:https://doi.org/10.3389/fpsyg.2012.00036

Martin, C., Bhui, R., and Bossaerts, P. (2014). Chimpanzee choice rates in competitive games match equilibrium game theory predictions. *Sci Rep* 4, 51–81

Miller, L. M. and D'Esposito, M. (2005). Perceptual fusion and stimulus coincidence in the cross-modal integration of speech. *Journal of Neuroscience* 25, 5884–5893

Monteiro, S. M. and Norman, G. (2013). Diagnostic reasoning: Where we've been, where we're going. *Teaching and Learning in Medicine* 25, S26–S3. doi:10.1080/10401334.2013.842911

Nivel, E., Thórisson, K. R., Steunebrink, B., Dindo, H., Pezzulo, G., Rodriguez, M., et al. (2013). Bounded recursive self-improvement. *Tech report RUTR-SCS13006, Reykjavik University – School of Computer Science*

Noesselt, T., Riegerand, J. W., Schoenfeld, M. A., Kanowski, M., Hinrichs, H., and Heinze, H. J. (2007). Audiovisual temporal correspondence modulates human multisensory superior temporal sulcus plus primary sensory cortices. *Journal of Neuroscience* 27, 11431–11441

Papaioannou, A. G., Kalantzi, E., Papageorgiou, C. C., and Korombili, K. (2021). Complexity analysis of the brain activity in autism spectrum disorder (asd) and attention deficit hyperactivity disorder (adhd) due to cognitive loads/demands induced by aristotle's type of syllogism/reasoning. a power spectral density and multiscale entropy (mse) analysis. *Heliyon* 7, e07984

Perry, J. C., Pakkenberg, B., and Vann, S. D. (2018). Striking reduction in neurons and glial cells in anterior thalamic nuclei of older patients with down's syndrome. *BioRxiv 449678* doi:doi:10.1101/449678

Sampathkumar, V., Miller-Hansen, A., Sherman, S. M., and Kasthuri, N. (2021). Integration of signals from different cortical areas in higher order thalamic neurons. *PNAS* 118, e2104137118. doi:10.1073/pnas.2104137118

Steenkiste, S. V., Chang, M., Greff, K., and Schmidhuber, J. (2018). Relational neural expectation maximization: Unsupervised discovery of objects and their interactions. *arXiv preprint*

Sumner, P., Tsai, P. C., Yu, K., and Nachev, P. (2006). Attentional modulation of sensorimotor processes in the absence of perceptual awareness. *PNAS* 103, 10520–10525

Thórisson, K. R. (2020). Seed-programmed autonomous general learning. In *Proceedings of Machine Learning Research*. 32–70

Tyll, S., Budinger, E., and Noesselt, T. (2011). Thalamic influences on multisensory integration. *Commun. Integr. Biol* 4, 145–171

Vaswani, A., Shazeer, N., Parmar, N., Uszkoreit, J., Jones, L., Kaiser, A. N. G. L., et al. (2017). Attention is all you need. In *Proc. Advances in Neural Information Processing Systems*. 5998–6008

Wang, P. (2005). Experience-grounded semantics: A theory for intelligent systems. *Cogn. Syst. Res.* 6, 282–302. doi:10.1016/j.cogsys.2004.08.003

Wang, P. (2006). *Rigid Flexibility: The Logic of Intelligence* (Dordrecht: Springer)

[Dataset] Wang, P. (2010). Non-axiomatic logic (nal) specification

Werner, S. and Noppeney, U. (2010). Superadditive responses in superior temporal sulcus predict audiovisual benefits in object categorization. *Cerebral Cortex* 20, 1829 – 1842

Wolff, M. and Vann, S. D. (2019). The cognitive thalamus as a gateway to mental representations. *J Neurosci* 39, 3–14. doi:10.1523/JNEUROSCI.0479-18

Yeh, C.-C. M., Zhu, Y., Ulanova, L., Begum, N., Ding, Y., Dau, A., et al. (2016). Matrix profile i: All pairs similarity joins for time series: A unifying view that includes motifs, discords and shapelets. 1317–1322. doi:10.1109/ICDM.2016.0179